\documentclass[runningheads]{llncs}
\usepackage[T1]{fontenc}
\usepackage{graphicx}
\usepackage{amsfonts}
\usepackage{amsmath}
\usepackage{multirow}
\usepackage{booktabs} 
\usepackage{siunitx}  

\usepackage{subcaption}

\usepackage[hidelinks,colorlinks=true,linkcolor=blue,citecolor=blue]{hyperref}

\begin{document}
\title{Large Intestine 3D Shape Refinement Using Conditional Latent Point Diffusion Models}

\titlerunning{3D Shape Refinement Using Latent Diffusion Models}
%
%
\author{Kaouther Mouheb\inst{1,2} \and
Mobina Ghojogh Nejad\inst{1} \and
Lavsen Dahal\inst{1,3} \and
Ehsan Samei\inst{1,3} \and
Kyle J. Lafata\inst{1,3} \and
W. Paul Segars\inst{1} \and
Joseph Y. Lo\thanks{Corresponding author: joseph.lo@duke.edu}\inst{1,3}}
\authorrunning{K. Mouheb et al.}
%
\institute{Center for Virtual Imaging Trials, Department of Radiology, Duke University School of Medicine, Durham, NC, USA \and
Biomedical Imaging Group Rotterdam, Department of Radiology \& Nuclear Medicine, Erasmus MC, Rotterdam, the Netherlands \and
Electrical and Computer Engineering, Pratt School of Engineering, Duke University, Durham, NC, USA
}
\maketitle              
\begin{abstract}
Accurate 3D modeling of human organs is critical for constructing digital phantoms in virtual imaging trials. However, organs such as the large intestine remain particularly challenging due to their complex geometry and shape variability. We propose CLAP, a novel \textbf{C}onditional \textbf{LA}tent \textbf{P}oint-diffusion model that combines geometric deep learning with denoising diffusion models to enhance 3D representations of the large intestine. Given point clouds sampled from segmentation masks, we employ a hierarchical variational autoencoder to learn both global and local latent shape representations. Two conditional diffusion models operate within this latent space to refine the organ shape. A pretrained surface reconstruction model is then used to convert the refined point clouds into meshes. CLAP achieves substantial improvements in shape modeling accuracy, reducing Chamfer distance by 26\% and Hausdorff distance by 36\% relative to the initial suboptimal shapes. This approach offers a robust and extensible solution for high-fidelity organ modeling, with potential applicability to a wide range of anatomical structures.

\end{abstract}
\section{Introduction}
In virtual imaging trials, realistic digital phantoms are crucial because they serve as virtual patients. Together with simulated image acquisition and interpretation, they allow researchers to conduct rigorous experiments with myriad parameter combinations, while reducing concerns about the high costs of trials and excessive radiation exposure \cite{segars2013population,hesterman2017three,wang2016novel}. These phantoms are typically created by segmenting organs from medical images and converting them into deformable mathematical representations such as polygon meshes \cite{segars20104d}. While early approaches relied on manual segmentation or traditional image processing methods \cite{segars20104d}, recent deep learning (DL) models such as TotalSegmentator have substantially advanced multi-organ segmentation from CT scans \cite{wasserthal2022totalsegmentator}. Despite these improvements, accurate segmentation of several anatomical structures remains a challenge \cite{cerrolaza2019computational}.
Segmenting the large intestine is particularly difficult due to its complex and variable shape, heterogeneous filling, and low contrast with surrounding soft tissue organs. Common problems include disconnected surfaces (over-segmentation) and the inclusion of nearby organs (under-segmentation) \cite{wang2022bowelnet}. Current DL methods like U-Nets \cite{ronneberger2015u} use voxel-based CNNs that lack shape awareness, leading to inaccurate surface reconstructions \cite{yang2022implicitatlas,raju2022deep}. Manually correcting the resulting shapes is labor intensive and subjective, underscoring the need for automated methods that ensure anatomically plausible shapes.

3D shape refinement has been extensively studied in domains such as autonomous driving and robotics, where LiDAR sensors produce sparse and noisy point clouds \cite{bai2020depthnet,varley2017shape}. Geometric deep networks, such as PointNet \cite{qi2017pointnet}, have been developed to process point clouds effectively. Denoising diffusion probabilistic models (DDPMs) \cite{ho2020denoising} have recently shown promise in completing missing data across different modalities. For example, the latent Point Diffusion Model (LION) employed a variational autoencoder (VAE) with a hierarchical latent space to capture shape distributions and generate high-quality shapes \cite{zeng2022lion}. For shape completion, the Point Diffusion-Refinement (PDR) approach consisted of a Conditional Generation Network (CGNet) and a Refinement Network (RFNet). CGNet is a DDPM that generates a coarse complete point cloud from partial input, while RFNet densifies the point cloud for quality improvement. Both networks utilize a dual path architecture based on PointNet++ for efficient feature extraction \cite{lyu2021conditional}. In medical imaging, similar methods have been used; for instance, Balsiger et al. used PointCNN to refine peripheral nervous system segmentations by classifying each point as foreground or background \cite{balsiger2019learning}. However, this method cannot recover false negatives missed by initial segmentation. Friedrich et al. applied point diffusion models to generate implants for defective skull shapes \cite{friedrich2023point}. Although effective in generating missing points, this approach cannot remove false positives, as the initial shapes remain fixed during the diffusion process. Recently, MedShapeNet was introduced as the first foundation model for medical point cloud completion \cite{yassin2024medshapenet}. Trained on the large-scale MedShapeNet dataset (100k samples), the model integrates 3D point clouds with text data to enhance shape reconstruction. However, class imbalance in the dataset may hinder refinement performance for underrepresented structures, particularly complex ones.

In this work, we address the challenge of inaccurate shape modeling of the large intestine caused by volumetric CT segmentation models. 
Inspired by LION and PDR, we propose a novel point cloud refinement method based on conditional point-diffusion models operating in a hierarchical latent space. We represent the organ shapes as point clouds and employ latent denoising diffusion models with PointNet-based backbones to predict complete and accurate reconstructions starting from the suboptimal shapes. Despite the small sample size of the dataset, our approach reconstructs anatomically plausible shapes, enabling both the generation of missing parts and the removal of false positives.

\section{Dataset}
We assembled a dataset of paired 3D large intestine shapes, consisting of suboptimal shape and the corresponding reference standard, drawn from two sources:\\
\textbf{TotalSegmentator dataset:}
We used the public TotalSegmentator dataset of 1,024 CT scans with reference segmentations of the large intestine \cite{wasserthal2022totalsegmentator}. To ensure complete coverage of the organ, we excluded scans with stomach volume below 50 ml or bladder volume equal to 0 ml (thereby ensuring the upper and lower limits of the intestine are present). Outliers with large intestine volumes outside the 400-1500 ml range are removed. Binary closing was used to merge fragmented segmentations, and only scans with two components (intestine and background) were retained. This process yielded 308 valid cases.\\
\textbf{Duke Health dataset:}
The use of this retrospective dataset was deemed exempt by the Duke University Health System Institutional Review Board and conducted in compliance with the Health Insurance Portability and Accountability Act (HIPAA). The private dataset includes 381 Duke Health CT scans, comprising 112 PET/CT and 269 chest-abdomen-pelvis (CAP) studies. Large intestine segmentations were generated using the pretrained TotalSegmentator model and refined through binary closing and component filtering. Additionally, 34 CAP cases were manually corrected by a physician. After filtering out cases with more than two components, 270 cases remained. To evaluate real-world performance, we selected 20 additional cases (8 PET/CT, 12 CAP) exhibiting visible segmentation defects. These cases were not used during model development. A physician manually corrected the TotalSegmentator-generated masks to serve as ground-truth references. Cases with major anatomical abnormalities were excluded due to time constraints.\\
\textbf{Partial shape synthesis:}
To generate training pairs for supervised refinement, we created suboptimal input shapes paired with accurate target shapes. Using our curated dataset, we simulated realistic segmentation errors by training an underfitted 3D full-resolution nnU-Net \cite{isensee2021nnu} on 30 randomly-selected cases for 30 epochs. This model was then used to produce suboptimal masks for all scans.\\
\textbf{Point cloud extraction:}
Given the segmentation masks, the marching cubes algorithm \cite{lorensen1987marching} was used to extract organ surfaces represented as polygon meshes. Using Poisson disk sampling \cite{yuksel2015sample}, point clouds of 2048 points were extracted from the surface meshes. Finally, the dataset was globally standardized using the mean and standard deviation calculated over all shapes in the training set.
We divide the data into 405 cases for training, 61 for validation, and 112 for testing.

\section{Method}
\subsection{Problem formulation}
We formulate the task of 3D shape refinement as a conditional point cloud generation. A point cloud sampled from the surface of a segmentation mask is represented by N points with xyz coordinates in the 3D space: $x \in \mathbb{R}^{N\times 3}$. We assume the dataset $\mathcal{D}$ is composed of M data pairs ${\{(x^i,c^i )|1 \leq i \leq M\}}$, where $x^i$ is the $i^{th}$ reference point cloud and $c^i$ is the corresponding suboptimal point cloud. The goal is to create a conditional model that generates a correct shape $y^i$ that represents an anatomically plausible representation of a large intestine, using the input $c^i$ as a conditioner. Note that the generated shape $y^i$ is as close as possible, but does not necessarily match the reference shape $x^i$.
\begin{figure}[t]
\centering
  \includegraphics[width=\textwidth]{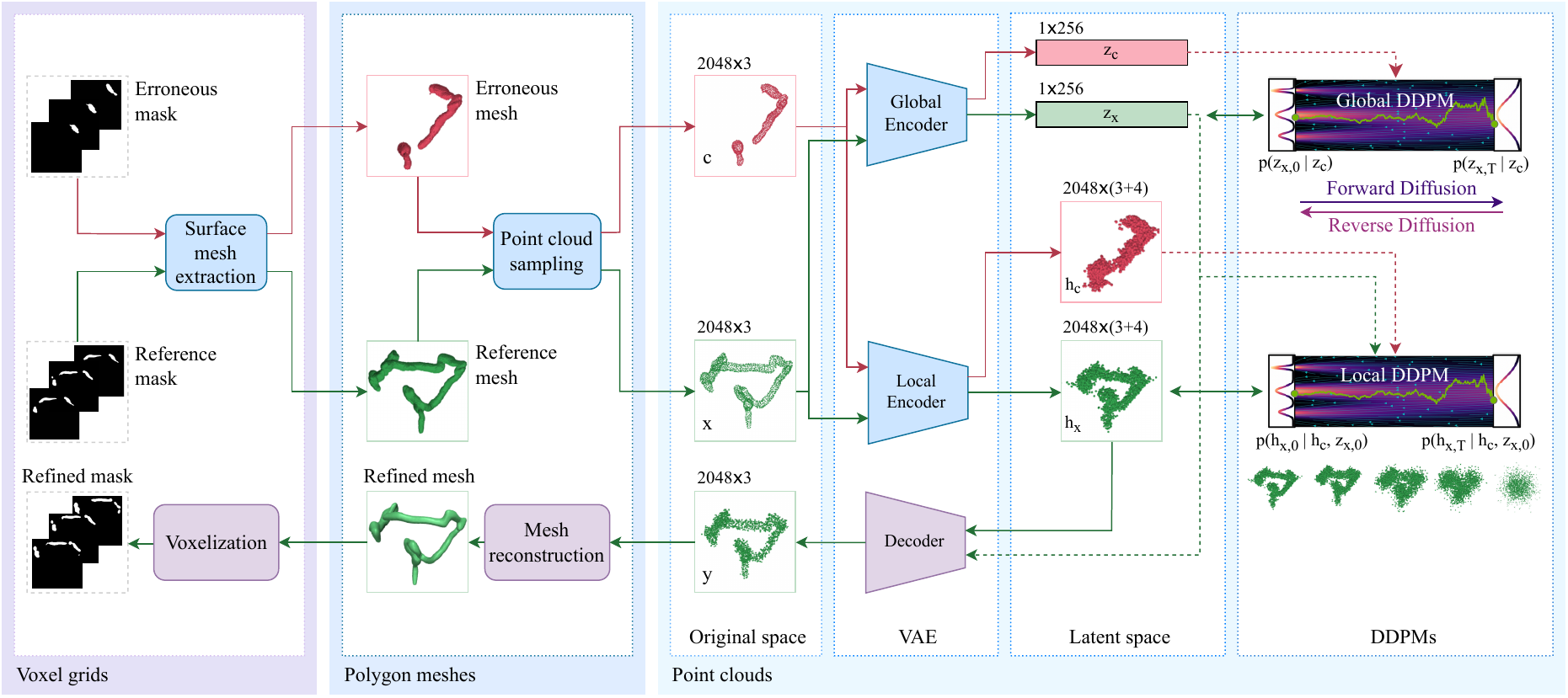}{}
   \caption{CLAP: Surface meshes are extracted from reference and suboptimal organ masks using the marching cubes algorithm, then converted to point clouds via Poisson disk sampling. A dual-encoder VAE maps each shape to global and local latent representations. Two conditional DDPMs are trained to model the latent distributions of complete shapes, conditioned on the latent representations of the suboptimal shapes. The refined shape is reconstructed via the VAE decoder. Solid arrows indicate inputs; dashed arrows indicate conditioning.}
   \label{fig:framework}
\end{figure}
\subsection{CLAP: Conditional latent point diffusion for shape refinement}
The proposed framework consists of two main stages. First, a VAE encodes each shape into a hierarchical latent space, producing both global and local latent representations. Second, two conditional DDPMs are trained to generate the latents of the correct shapes, conditioned on the latents of the suboptimal shapes. The motivation behind shifting the diffusion model training into a smoother latent space lies in the high complexity and variability of the large intestine's shapes, making it challenging for the DDPM to accurately model their distribution in the original space. An overview of the framework is shown in Fig.~\ref{fig:framework}.
\subsubsection{Hierarchical latent shape encoding:}
We adapt the VAE of LION \cite{zeng2022lion} to encode shapes into a hierarchical latent space. The VAE consists of two encoders and one decoder based on the PVCNN architecture \cite{liu2019point}. For each training pair $(x, c) \in \mathcal{D}$, the first encoder extracts a global latent vector \( z_s \in \mathbb{R}^{D_z} \) for each shape \( s \in \{x,c\} \). The second encoder takes the shape $s$ as input and the latent $z_s$ as condition to extract a local latent point cloud \( h_s \in \mathbb{R}^{N\times(3+D_h)} \), where each of the \( N \) points carries its spatial coordinates $(x, y, z)$ and \( D_h \) additional features. The decoder takes the latent representations $z_s$ and $h_s$ as input to reconstruct the shape $s$ in the original space. The VAE is trained by maximizing a modified Evidence Lower Bound (ELBO) loss that includes a reconstruction term and two Kullback–Leibler (KL) divergence terms. The loss with respect to the encoder's parameters $\theta$ and the decoder's parameters $\psi$  is defined as:
\begin{align}
\mathcal{L}(\theta, \psi) =\; & \mathbb{E}_{q_\theta(z_s, h_s|s)}\Big[ \log p_\psi(s|z_s, h_s) \nonumber \\
& - \lambda_z\, \text{KL}\left(q_\theta(z_s|s)\,\Vert\,p(z_s)\right) 
- \lambda_h\, \text{KL}\left(q_\theta(h_s|s,z_s)\,\Vert\,p(h_s)\right) \Big]
\end{align}
where \( \lambda_z \) and \( \lambda_h \) control the influence of KL regularization for the global and local latents, respectively. The KL weights are gradually annealed to their maximum values, encouraging the latent priors \( p(z_s) \) and \( p(h_s) \) to converge to standard Gaussian distributions. The hierarchical nature of the latent space helps to disentagle the high-level shape features from the fine-grained details which further simplifies the distributions to be learned by the diffusion models therefore reducing the number of data points required for training.
\subsubsection{Conditional latent point diffusion:}
DDPMs learn to reverse a gradual noising process applied to data over multiple time steps \cite{ho2020denoising}. Starting from pure Gaussian noise, a trained DDPM denoises the input step-by-step using a neural network. When conditioned on auxiliary inputs, this framework enables controlled generation. Appendix A provides a mathematical formulation of DDPMs.

In CLAP, two DDPMs are trained in the latent space learned by the VAE. The first DDPM, parameterized by $\xi$, models the global latent space. It learns to denoise the global latent $z_x$ of the reference shape, conditioned on the global latent $z_c$ of the suboptimal shape. The second DDPM, parameterized by $\phi$, operates on the local latent representations. It denoises $h_x$ conditioned on both $h_c$ and $z_x$. Both models are trained by minimizing the difference between the actual noise added to the reference latent encodings and the noise predicted by the DDPM. The training loss of the global DDPM can be written as:
\begin{align}
    \mathcal{L}(\xi) = \mathbb{E}_{i \sim \mathcal{U}([M]),\, t \sim \mathcal{U}([T]),\, \epsilon \sim \mathcal{N}(0, I)} 
    \left\| \epsilon - \epsilon_\xi(z_{x,t}^i,\, t,\, z_c^i) \right\|^2
\end{align}
where $\mathcal{U}([M])$ represents the uniform distribution over \{1, 2,...,M\}, $\mathcal{U}(T)$ represents the uniform distribution over \{1, 2,...,T\} with $T$ being the maximum number of diffusion steps, $z_{x,t}^i$ is the diffused global latent representation of the shape $x^i$ after $t$ diffusion steps, $z_c^i$ is the  global representation of the corresponding conditioner $c^i$, $\epsilon$ is the actual noise and $\epsilon_\xi$ is the noise predicted by the model. Similarly, the loss of the local DDPM can be written as:
\begin{align}
    \mathcal{L}(\phi) = \mathbb{E}_{i \sim \mathcal{U}([M]),\, t \sim \mathcal{U}([T]),\, \epsilon \sim \mathcal{N}(0, I)} 
    \left\| \epsilon - \epsilon_\phi(h_{x,t}^i,\, t,\, h_c^i,\, z_{x,0}^i) \right\|^2
\end{align}
where $h_{x,t}^i$ is the diffused local latent representation of the shape $x^i$ after t diffusion steps, $h_c^i$ is the local representation of the corresponding conditioner $c^i$, $z_{x,0}^i$ is the clean global representation of $x^i$, $\epsilon$ is the actual noise and $\epsilon_\phi$ is the noise predicted by the model.

The global DDPM is implemented as a ResNet with eight squeeze-and-excitation blocks. The local DDPM adopts a dual-path PointNet-based architecture from the CGNet model \cite{lyu2021conditional}. In both models, the conditioning latents $z_c$ and $h_c$ (from the suboptimal shapes) are not diffused. Instead, features are extracted from these latents and fused into the denoising network to guide the generation process. This conditioning enables the model to learn both to add missing structures and to suppress incorrect regions, effectively refining the suboptimal shape toward a more anatomically accurate form.
\subsubsection{Inference:}
\begin{figure}[t]
\centering
  \includegraphics[width=\textwidth]{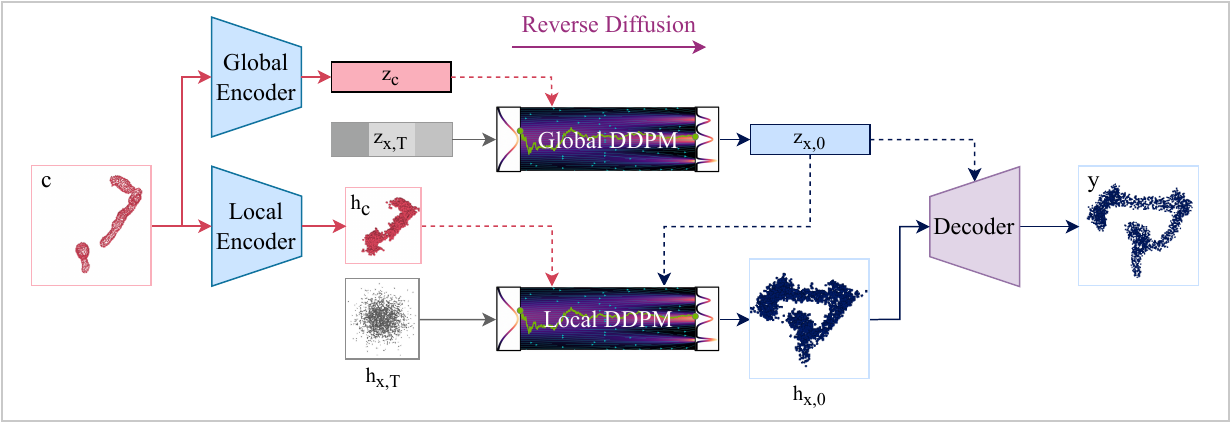}{}
   \caption{Inference pipeline:  The erroneous shape c is encoded into latent representations $(z_c, h_c)$, and noisy inputs $z_{x, T}$ and $h_{x, T}$ are sampled. Reverse diffusion processes are applied using the global and local representations of c as conditions to obtain clean representations $z_{x, 0}$ and $h_{x, 0}$. These representations are then decoded back to the original space using the VAE's decoder.}
   \label{fig:inference}
\end{figure}

In inference, the suboptimal shape $c$ is encoded to its latent representations $(z_c, h_c)$, and two noisy inputs $z_{x, T}$ and $h_{x, T}$ are sampled from a normal Gaussian distribution. First, the reverse diffusion process is performed on $z_{x, T}$ using the global DDPM to obtain a global representation $z_{x, 0}$. Later, the local DDPM is used to run the reverse diffusion process on $h_{x, T}$ to obtain a refined local representation $h_{x, 0}$ using both $h_c$ and $z_{x, 0}$ as conditions. Finally, the resulting representations are decoded back to the original space using the VAE's decoder. 
This inference process is depicted in Fig. \ref{fig:inference}.
\subsubsection{Post-processing and surface reconstruction:}
The generated point clouds can be noisy and sparse, requiring post-processing before surface reconstruction. First, point clouds are rescaled to their original dimensions and smoothed using moving least squares. Densification is performed by inserting midpoints between distant neighbors (over 10 mm apart within a 10-point neighborhood), then outliers with fewer than five neighbors within a 15 mm radius are removed.
Finally, we reconstruct the 3D mesh using the open-source Transformer-based model for point cloud-to-mesh reconstruction from the Point-E framework \cite{nichol2022point}.
\subsubsection{Hyperparameters and implementation details:}
The number of input and output point clouds is set to 2048. The maximum values of the KL weights of the loss function $\lambda_z$ and $\lambda_h$ were set to 0.4. The dimension of the global latents is set to $D_z=256$ and the number of features of the local latent points is set to $D_h=4$.  The VAE was trained for 6000 epochs using the Adam optimizer with a batch size of 32 and a learning rate of $10^{-3}$. The DDPMs were trained in parallel using Adam optimizer with $T=1000$ diffusion steps, a batch size of 10 and a learning rate of $2\times10^{-4}$ for 16000 epochs. The code is publicly available at \url{https://gitlab.com/kaouther.mouheb/clap}

\section{Results}
We compare the performance of CLAP to the pretrained MedShapeNet model \cite{yassin2024medshapenet}, which we used to generate refined shapes for the test set by providing “colon” as input to its text encoder, and the suboptimal shape as the partial point cloud. Besides, we trained CGNet on our dataset using the code provided by the authors \cite{lyu2021conditional}. Shape refinement was evaluated using the Chamfer distance (CD) and the Hausdorff distance (HD) between the generated shapes and the reference standard \cite{pcu}. Wilcoxon signed-rank test was used for statistical testing.

\subsection{3D Shape Refinement}
\begin{figure}[t]
\centering
  \includegraphics[width=\textwidth]{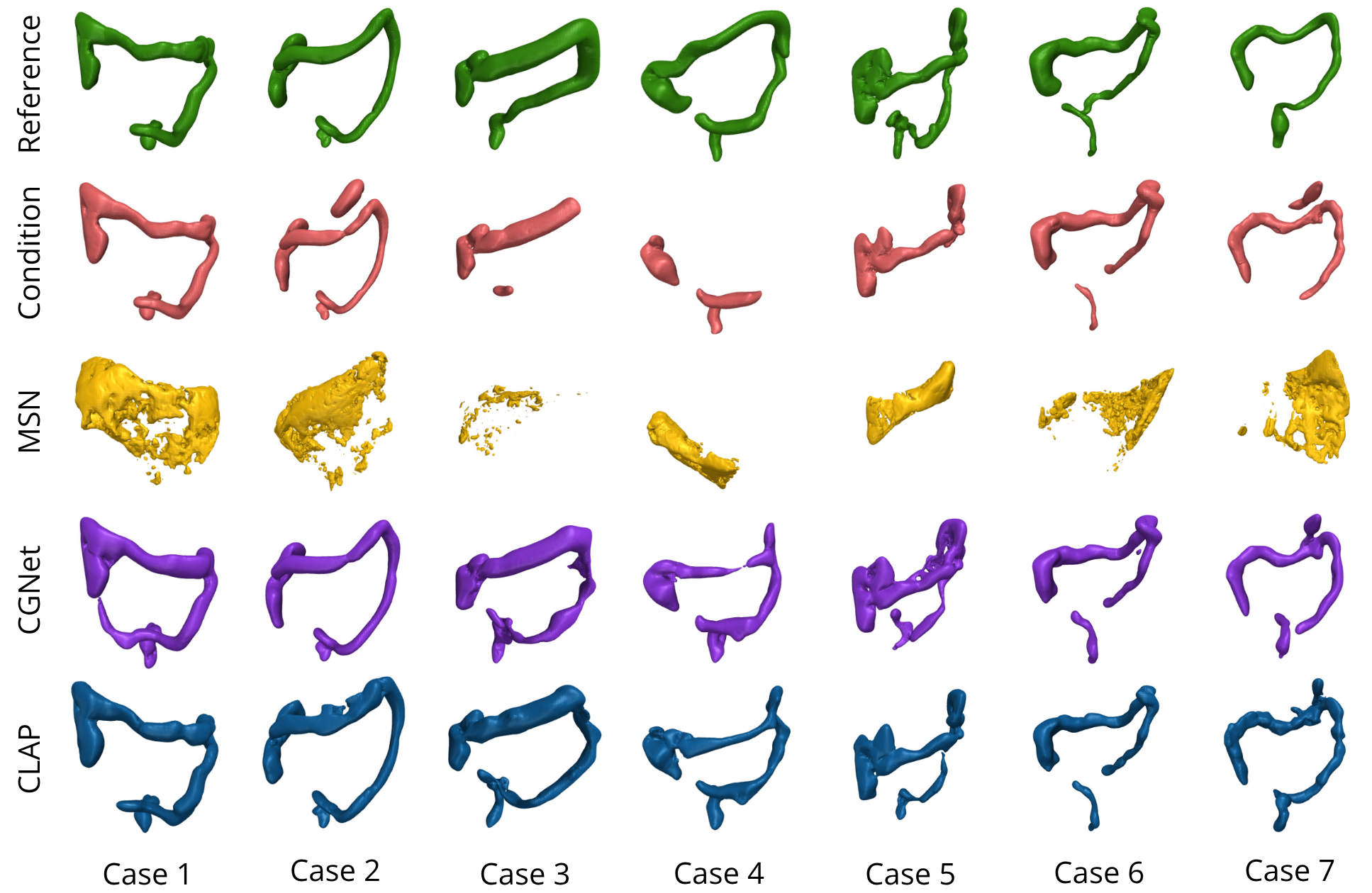}{}
   \caption{Example qualitative results illustrating shape refinement performance. The first row displays the reference shapes. The second row presents the suboptimal shapes used as input to the models. The third to fifth rows show the refined outputs produced by MedShapeNet (MSN), CGNet, and CLAP, respectively.}
   \label{fig:examples}
\end{figure}

Table \ref{tab:quantitative_results} shows the quantitative results for each method. We consider cases where the initial CD between the suboptimal shape and the reference is less than 10 mm as easy cases (n = 55) and those with a CD greater than 10 mm as hard cases (n = 58). Qualitative examples are shown in Fig. \ref{fig:examples}.

\begin{table}[t]
\centering
\caption{Quantitative comparison of shape refinement performance using Chamfer Distance (CD) and Hausdorff Distance (HD). Init refers to the distance between the suboptimal input shapes and the physician-annotated reference. MSN denotes MedShapeNet.}
\label{tab:quantitative_results}
\footnotesize
\setlength{\tabcolsep}{3pt}
\begin{tabular}{@{}l*{6}{c}@{}}
\toprule
\multirow{2}{*}{Method} & \multicolumn{3}{c}{CD (mm) $\downarrow$} & \multicolumn{3}{c}{HD (mm) $\downarrow$} \\ 
\cmidrule(lr){2-4}\cmidrule(lr){5-7}
 & Easy & Hard & All & Easy & Hard & All \\ 
\midrule
Init & \textbf{5.6{\scriptsize $\pm$2.3}} & 30.2 {\scriptsize $\pm$21.2} & 17.8 {\scriptsize $\pm$19.4} & 45.1 {\scriptsize $\pm$21.2} & 128.9 {\scriptsize $\pm$50.1} & 86.6 {\scriptsize $\pm$56.8} \\
MSN & 21.4{\scriptsize $\pm$7.6} & 29.5{\scriptsize $\pm$10.7} & 25.5{\scriptsize $\pm$10.1} & 74.4{\scriptsize $\pm$14.5} & 100.7{\scriptsize $\pm$28.4} & 87.9{\scriptsize $\pm$26.2} \\ 
CGNet & 6.4 {\scriptsize $\pm$2.6} & 20.5 {\scriptsize $\pm$12.1} & 13.3 {\scriptsize $\pm$11.2} & 42.4 {\scriptsize $\pm$18.6} & 75.6 {\scriptsize $\pm$33.4} & 58.7 {\scriptsize $\pm$31.5} \\ 
CLAP & 6.8 {\scriptsize $\pm$2.3} & \textbf{19.8 {\scriptsize $\pm$12.0}} & \textbf{13.2 {\scriptsize $\pm$10.7}} & \textbf{40.4 {\scriptsize $\pm$18.6}} & \textbf{71.6 {\scriptsize $\pm$34.8}} &  \textbf{55.7 {\scriptsize $\pm$31.8}} \\ 
\bottomrule
\end{tabular}
\end{table}

The initial suboptimal shapes serve as a baseline, showing low CD for easy cases (5.6 ± 2.3 mm), but drastically higher errors on hard cases (30.2 ± 21.2 mm) and overall (17.8 ± 19.4 mm). Similarly, HD values are notably high in all categories, especially for hard cases (128.9 ± 50.1 mm), indicating large outlier deviations. MedShapeNet significantly underperforms compared to the other methods (\textit{p}<0.001), with overall CD of 25.5 ± 10.1 mm and HD of 87.9 ± 26.2 mm. Both CGNet and CLAP significantly improve performance compared to the initial shapes, particularly in terms of HD (\textit{p}<0.001). CGNet achieves a marked reduction in both CD and HD (13.3 ± 11.2 mm and 58.7 ± 31.5 mm overall), with strong performance, particularly in easy cases. However, CLAP outperforms all methods, achieving the lowest overall CD (13.2 ± 10.7 mm) and HD (55.7 ± 31.8 mm, significantly lower than CGNet with \textit{p}<0.02), and demonstrating consistent improvement in both easy and hard scenarios. Notably, CLAP also yields the best results for the most challenging cases in terms of both CD (19.8 ± 12.0 mm) and HD (71.6 ± 34.8 mm), suggesting its superior generalization capability and robustness to shape complexity. Qualitative evaluation shows that CLAP successfully preserves correct shapes (Case 1), removes false positives (Case 2), and completes partial structures (Case 3) even in challenging cases with significant missing regions (Cases 4, 5). While CGNet produces accurate shapes, it occasionally introduces anatomical errors by adding extraneous parts (Cases 1, 5) or failing to connect fragments (Case 4). CLAP fails to remove attached false positives in some cases, but often merges them into a more plausible connected structure (Case 7). In some instances, it fails to reconnect organ segments (Case 6). Notably, MedShapeNet fails to refine the shapes for all cases.

\subsection{TotalSegmentator Refinement}
\begin{figure}[t]
\centering
\begin{minipage}[t]{0.48\textwidth}
  \centering
  \includegraphics[width=\textwidth]{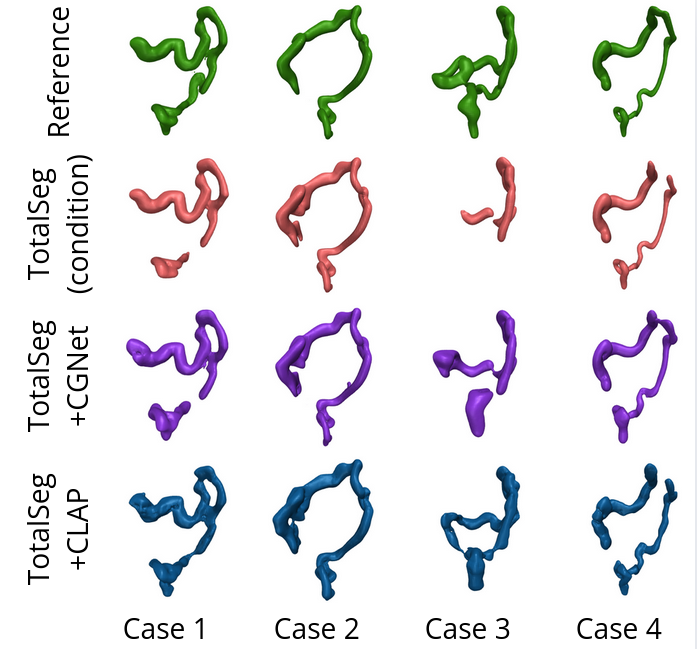}
  \caption{Example shape refinement results on real-world cases. Row 1: physician-refined reference shapes. Row 2: suboptimal shapes from TotalSegmentator. Rows 3–4: outputs from CGNet and CLAP, respectively.}
  \label{fig:totalseg}
\end{minipage}
\hfill
\begin{minipage}[t]{0.48\textwidth}
  \centering
  \includegraphics[width=\textwidth]{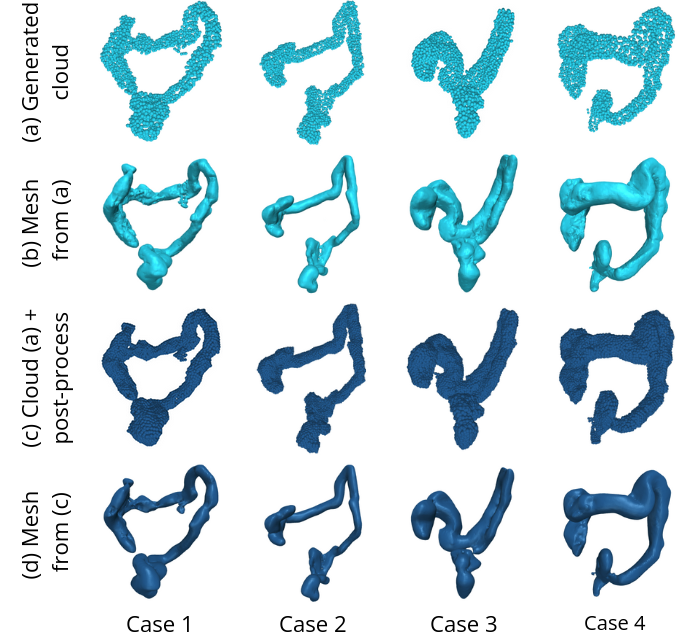}
  \caption{Impact of post-processing on CLAP outputs. (a) Raw point clouds from CLAP, (b) meshes obtained from raw clouds, (c) point clouds after post-processing, (d) meshes obtained from the post-processed clouds.}
  \label{fig:postprocess}
\end{minipage}
\end{figure}

We evaluated our pipeline on 20 real-world Duke Health cases with mild to moderate segmentation defects. These cases were selected and manually corrected by a physician to serve as reference standards (see Methods). Visual results are shown in Fig.~\ref{fig:totalseg}. MedShapeNet was not included in this experiment due to its previously observed poor performance.

The initial shapes obtained from the segmentation masks generated by TotalSegmentator yielded a mean CD of 23.2 ± 43.7 and a mean HD of 115.1 ± 153.9, indicating substantial deviations from the physician's reference. CGNet refinement reduced the discrepancies, with a CD of 11.1 ± 9.7 and HD of 64.0 ± 63.0. The best performance was achieved using CLAP, which improved alignment with the reference annotations, achieving the lowest CD of \textbf{9.9 ± 6.2} and HD of \textbf{54.8 ± 60.1}. These results demonstrate a consistent improvement in 3D shape modeling, with CLAP providing the most precise correspondence to expert delineations. The qualitative evaluation confirms that the observations made on the synthetic set (accurate distribution modeling, noise elimination in certain cases, and failure scenarios) are consistent with real-world cases.

\subsection{Computational Time Analysis}
The VAE's training time was 38 hours distributed on 2 NVIDIA RTX A6000 GPUs whereas the training of the DDPMs lasted for 63 hours distributed on 4 NVIDIA RTX A6000 GPUs. We computed an estimation of the average inference time of each component of our refinement pipeline on a set of 20 samples. The results are shown in table \ref{tab:comp_time}.

\setlength{\tabcolsep}{20pt} 
\begin{table}[t]
\centering
\caption{Average computational time required for each process.}
\label{tab:comp_time}
\begin{tabular}{@{}l r@{\,±\,}l@{}}
\toprule
\textbf{Process} & \multicolumn{2}{c}{\textbf{Time (s)}} \\
\midrule
Marching cubes       &  4.39  & 3.5  \\
Poisson disk sampling&  0.09  & 0.0  \\
VAE+DDPM inference   & 96.83  & 1.8  \\
Post-processing      &  0.44  & 0.1  \\
Point-E pc2mesh      &  9.80  & 2.2  \\
\midrule
Complete pipeline    &111.56  & 4.5  \\
\bottomrule
\end{tabular}
\end{table}

On average, the entire pipeline can be executed in less than 2 minutes for a single case. It is worth noting that the latent DDPM sampling process consumed the most time, as expected, due to the execution of 1000 reverse diffusion steps for each of the two diffusion models. It is important to highlight that the process was individually applied to each sample. Processing the clouds in batches during the execution of the reverse diffusion processes and the mesh reconstruction could significantly decrease the average time for a larger set of samples.

\subsection{Ablation Experiments}

\textbf{Impact of the latent space distribution:} During VAE training, the KL divergence weights \( \lambda_z \) and \( \lambda_h \) are progressively increased, encouraging the approximate posteriors to align more closely with the standard Gaussian prior. As a result, the latent representations \( p(z_s) \) and \( p(h_s) \) exhibit reduced variance collapse and better coverage of the latent space, with distributions that are more isotropic and closer to a unit Gaussian in both mean and covariance. While this improves latent space regularization and facilitates sampling, it increases the reconstruction error. To assess the effect of the latent space's distribution on refinement quality, three latent DDPMs were trained using VAE checkpoints saved at epochs 4000, 6000, and 8000. Table~\ref{tab:ablation_vae} reports the VAE's reconstruction performance on the test set using the F1 score, following the metric proposed by Tatarchenko et al.~\cite{Tatarchenko_2019_CVPR}, as well as the refinement performance of the corresponding DDPMs in terms of CD and HD.

The results show that with longer training, the VAE's reconstruction performance slightly decreases (F1 Score of 98.6 at 4000 epochs vs. 98.3 at 8000 epochs). The best refinement performance, in terms of both CD and HD, is achieved at 6000 training epochs (CD: 13.2 ± 10.7, HD: 55.7 ± 31.8) yielding the optimal balance between the reconstruction performance of the VAE and downstream shape refinement.\\
\\
\textbf{Point cloud post-processing:}
To evaluate the impact of the proposed post-processing pipeline, we generated meshes from both the raw and post-processed point clouds. Figure~\ref{fig:postprocess} illustrates representative examples of the resulting point clouds and the corresponding meshes. Quantitatively, the raw point clouds yielded meshes with a slightly lower CD of 13.1 ± 9.7 mm compared to 13.2 ± 10.7 mm for those generated from post-processed clouds. However, the corresponding HD was higher (58.3 ± 30.4 mm vs. 55.7 ± 31.8 mm), reflecting a greater number of outlier points in the raw point clouds. Qualitatively, post-processing leads to cleaner and smoother mesh surfaces with fewer holes and artifacts, aligning better with the anatomical structure of the large intestine's wall, which consists of soft tissue.

\begin{table}[t]
\centering
\caption{Ablation study on VAE training epochs showing reconstruction performance (F1 score) and point shape refinement metrics (CD and HD).}
\label{tab:ablation_vae}
\setlength{\tabcolsep}{10pt} 
\begin{tabular}{lccc}
\toprule
VAE Training & Reconstruction & \multicolumn{2}{c}{Refinement Metrics} \\
\cmidrule(lr){2-2} \cmidrule(lr){3-4}
Epochs & F1 (\%) $\uparrow$ & CD (mm) $\downarrow$ & HD (mm) $\downarrow$ \\
\midrule
4000 & 98.6{\scriptsize $\pm$0.6} & 13.4{\scriptsize $\pm$10.5} & 57.9{\scriptsize $\pm$31.2} \\
6000 & 98.5{\scriptsize $\pm$0.6} & 13.2{\scriptsize $\pm$10.7} & 55.7{\scriptsize $\pm$31.8} \\
8000 & 98.3{\scriptsize $\pm$0.7} & 13.8{\scriptsize $\pm$10.7}  & 58.3{\scriptsize $\pm$32.4} \\
\bottomrule
\end{tabular}
\end{table}
\section{Discussion}
In this work, we used geometric deep learning and denoising diffusion models to refine 3D shapes of the large intestine obtained from CT segmentation models. Our approach framed this task as a conditional point cloud generation problem. We used a hierarchical VAE to encode shapes into a smoother latent space with global and local representations. Two denoising diffusion models refined the point clouds in this latent space before decoding back to the original space.

Our findings highlight the limitations of general-purpose foundation models in this task, as we showed that MedShapeNet fails to refine the shapes of the large intestine. This could be due to the limited number of samples of the intestine class (named ``colon'' in MedShapeNet) in its training data (1\% of the total set), which can make the model biased towards other structures.
Compared to CGNet, our method showed that training denoising diffusion models in a hierarchical latent space improved the modeling of shape distributions, particularly in capturing fine details. Our method effectively handled false positives, while CGNet sometimes produced anatomically inaccurate details, such as introducing small branches into the colon. 
The study on the impact of latent space smoothness showed that models trained in smoother spaces preserved the organ's anatomy more accurately. We additionally show the effectiveness of our proposed post-processing pipeline in improving the smoothness of the resulting surfaces, reflecting a higher fidelity to the anatomy of the organ. We applied our pipeline to the outputs produced by TotalSegmentator, a widely recognized model trained on one of the largest public CT datasets. This resulted in notable improvements in surface representation performance. Furthermore, qualitative evaluation confirmed the consistency of our results with those obtained on the synthetic shapes, demonstrating the fidelity of our synthetic dataset to real segmentation outputs and the ability of our method to generalize to unseen cases.

Although our method demonstrates promise for large intestine shape refinement, it has a number of limitations. The model faces difficulties in eliminating false positives closely adjacent or attached to the actual segments of the organ. In addition, it is still challenging to connect large gaps between segments, especially near the rectosigmoid junction, and may produce anatomically inaccurate shapes for complex inputs. These errors are expected since the size of the dataset used to train the model is very small compared to common computer vision datasets typically used for this task such as ShapeNet \cite{Chang2015ShapeNetAI} which includes over 50000 shapes. Increasing the training set size and pretraining the model on larger sets may address these issues. Moreover, the generated shapes sometimes misalign with the reference standard and intersect with neighboring organs, which impacts the ability to create consistent computational phantoms with multiple organs. The current model relies only on the provided partial shape, lacking contextual information about the surrounding anatomy.

For future improvements, our goal is to increase the diversity of the data by including more samples and using data augmentation. Future efforts could also explore using landmarks from neighboring organs to condition the generative models to avoid intersections and improve shape accuracy, as well as optimizing the landmark extraction technique for a more generalizable synthesis approach. Ultimately, the goal is to generate new organs from scratch without the need for an initial segmentation. Finally, mesh reconstruction could be enhanced by fine-tuning the Point-E model or modifying the VAE decoder to output more advanced shape representations rather than point clouds (e.g. signed distance functions). These steps aim to eliminate false positives, diversify the dataset, and improve mesh reconstruction quality.

\section{Conclusion}
In this work, we introduced an end-to-end pipeline that enhances 3D surface reconstruction of the large intestine from imperfect segmentations. By integrating geometric deep learning with denoising diffusion probabilistic models, we tackled conditional point cloud generation in a hierarchical latent space, enabling anatomically faithful refinements. Our experimental results, both quantitative and qualitative, highlight notable improvements in reconstruction quality and anatomical accuracy. While the large intestine served as a representative case study, the proposed methodology demonstrates strong potential for generalization, paving the way for its application in constructing more realistic computational phantoms and advancing downstream biomedical modeling tasks.

\begin{credits}
\subsubsection{\ackname}
This work was supported by the National Institutes of Health under the following grant numbers: P41EB028744, R01EB001838, and R01CA261457. K. Mouheb was supported by the Erasmus Mundus Joint Master Degree program in Medical Imaging and Applications.
\subsubsection{\discintname}
The authors have no competing interests to declare that are
relevant to the content of this article.
\end{credits}

\bibliographystyle{splncs04}
\bibliography{references}
\end{document}